\pdfoutput=1

\documentclass[11pt]{article}

\usepackage[preprint]{acl}

\usepackage{times}
\usepackage{latexsym}

\usepackage[T1]{fontenc}

\usepackage[utf8]{inputenc}

\usepackage{microtype}
\usepackage{inconsolata}

\usepackage{graphicx}

\usepackage[whole]{bxcjkjatype}
\usepackage{amssymb}
\usepackage{amsmath}
\usepackage[percent]{overpic}
\usepackage{bm}

\usepackage{booktabs}
\usepackage{multirow} 
\usepackage{color}
\usepackage{lingmacros}
\usepackage{graphbox}
\usepackage{colortbl}

\newcommand{\gcella}{\cellcolor[rgb]{0.92, 0.95, 0.92}}
\newcommand{\gcellb}{\cellcolor[rgb]{0.8, 0.95, 0.8}}
\newcommand{\gcellc}{\cellcolor[rgb]{0.55, 0.95, 0.55}}

\usepackage[capitalise,english,nameinlink]{cleveref} 
\Crefname{figure}{\text{Figure}}{\text{Figures}}
\creflabelformat{equation}{#2\textup{#1}#3}
\crefname{line}{\text{line}}{\text{lines}} 
\crefname{assumption}{\text{Assumption}}{\text{Assumptions}} 
\usepackage{cite}

\ifdefined\VersionWithComments%
	\usepackage[colorinlistoftodos,textsize=footnotesize]{todonotes}
\else
	\usepackage[disable]{todonotes}
\fi

\setuptodonotes{inline}

\newlength{\needcitelength}
\newcommand{\needcite}[1]{\settowidth{\needcitelength}{cites: {#1}}\todo[noinlinepar,inlinewidth=\the\needcitelength,size=\scriptsize]{cites: {#1}}}
\newcommand{\needref}[1]{\settowidth{\needcitelength}{ref: {#1}}\todo[color=yellow,noinlinepar,inlinewidth=\the\needcitelength,size=\scriptsize]{ref: {#1}}}

\title{How a Bilingual LM Becomes Bilingual:\\Tracing Internal Representations with Sparse Autoencoders}

\author{
 \textbf{Tatsuro Inaba\textsuperscript{1}\thanks{Work completed as a research assistant at NII LLMC.}}\quad
 \textbf{Go Kamoda\textsuperscript{2,3}}\quad
 \textbf{Kentaro Inui\textsuperscript{1,4,5}}\quad
\\
 \textbf{Masaru Isonuma\textsuperscript{6,4,5}}\quad
 \textbf{Yusuke Miyao\textsuperscript{7,6}}\quad
 \textbf{Yohei Oseki\textsuperscript{7,6}}\quad
\\
 \textbf{Yu Takagi\textsuperscript{8}\thanks{Co-last authors}}\quad
 \textbf{Benjamin Heinzerling\textsuperscript{5,4}\footnotemark[2]}
\\
 \textsuperscript{1}MBZUAI\quad
 \textsuperscript{2}SOKENDAI\quad
 \textsuperscript{3}NINJAL\quad
 \textsuperscript{4}Tohoku University\quad
 \textsuperscript{5}RIKEN\quad
\\
 \textsuperscript{6}NII LLMC\quad
 \textsuperscript{7}University of Tokyo\quad
 \textsuperscript{8}Nagoya Institute of Technology
\\
 \small{
   \textbf{Correspondence:} \href{mailto:tatsuro.inaba@mbzuai.ac.ae}{tatsuro.inaba@mbzuai.ac.ae}
 }
}

\begin{document}
\maketitle
\begin{abstract}
This study explores how bilingual language models develop complex internal representations.
We employ sparse autoencoders to analyze internal representations of bilingual language models with a focus on the effects of training steps, layers, and model sizes.
Our analysis shows that language models first learn languages separately, and then gradually form bilingual alignments, particularly in the mid layers. 
We also found that this bilingual tendency is stronger in larger models.
Building on these findings, we demonstrate the critical role of bilingual representations in model performance by employing a novel method that integrates decomposed representations from a fully trained model into a mid-training model.
Our results provide insights into how language models acquire bilingual capabilities\footnote{\label{note1}Our code is publicly available at \url{https://github.com/llm-jp/llm-jp-sae}}.
\end{abstract}


\section{Introduction}
\label{sec:intro}
Large Language Models (LLMs) have demonstrated remarkable multilingual capabilities
~\citep{openai2024gpt4technicalreport,Dubey-2024-theLlama3HerdOfModels-vy,gemmateam2025gemma3technicalreport}.
However, it is not yet clear how such capabilities emerge during pre-training.
Specifically, do LLMs initially learn each language separately before aligning them?
Is cross-lingual alignment distributed across layers or concentrated in specific components?
How does model size affect this alignment process?
These are not just theoretical questions; they directly impact our understanding of model scalability and the emergence of generalization abilities~\citep{wei2022emergent}.

To address these questions, in this study, we explore the internal mechanisms through which LLMs develop their internal representations; namely, we trace when, where, and how bilingual alignment (English-Japanese) emerges during pretraining.
For this purpose, we use sparse autoencoders \citep[SAEs;][]{bricken2023monosemanticity, huben2024sparse} as a tool for our analysis, which enable us to extract interpretable latent features from hidden representations.
Unlike previous approaches \citep{bricken2023monosemanticity,huben2024sparse,balcells2024evolutionsaefeatureslayers,wang2025towards}, our method captures fine-grained distinctions between language-specific and bilingual features, as well as semantic features, and allows analysis of their emergence across training stages and model layers. 

\begin{figure}[t]
\centering
\includegraphics[width=\columnwidth]{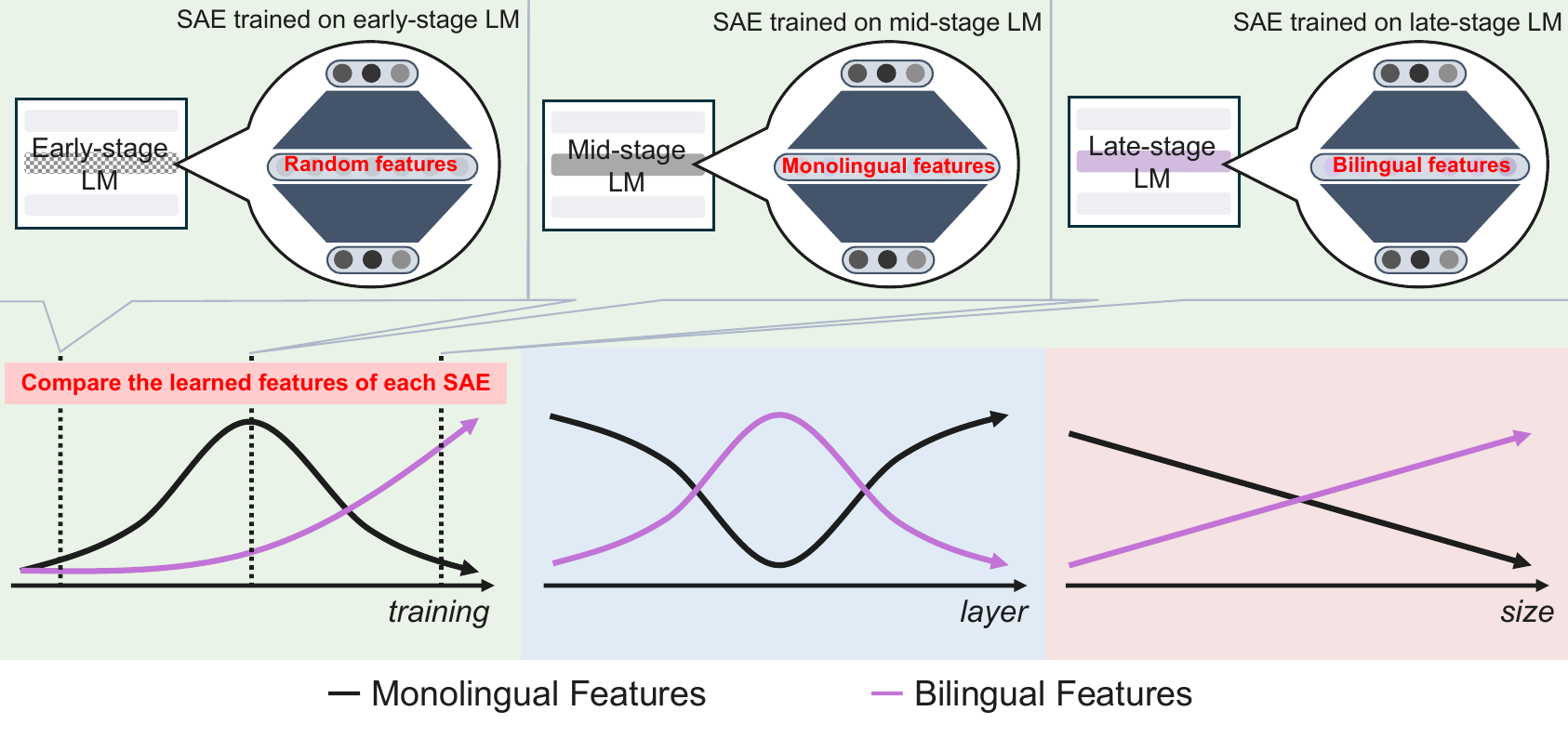}
\caption{Illustration of the experimental setup (top) and the key findings (bottom). 
In the top panel, SAEs are trained independently on language models at each training stage, layer, and model size. 
The bottom panel visualizes the evolution of bilingual alignment, derived from comparisons of the features learned by each SAE.}
\label{fig:overview}
\end{figure}

We conduct experiments on decoder-only models with a variety of sizes, pretrained on an English-Japanese bilingual corpus.
Our observations highlight three key findings, as illustrated in \cref{fig:overview}.
\begin{itemize}
    \item LLMs initially learn languages independently, and gradually develop bilingual alignment over training (\cref{sec:ckpt}).
    \item Bilingual alignments are more prominently captured in the mid-layers of the model (\cref{sec:layers}).
    \item Larger models exhibit stronger bilingual alignment than smaller ones (\cref{sec:sizes}).
\end{itemize}

Beyond these observations, we introduce an SAE-based method to identify which types of representations are most essential to the model.
We first decompose the representations of a fully trained model into three distinct types: English-specific, Japanese-specific, and bilingual.
These components are then selectively injected into the model at a mid-training stage, allowing us to evaluate their importance by analyzing the resulting changes in the model's behavior.

Our results demonstrate that bilingual representations from a fully trained model enhance the performance of a mid-training model (\cref{sec:add_bi}).
Beyond simply using SAEs to interpret language models, we harnessed them to directly manipulate internal representations, demonstrating their versatility as tools for both analysis and intervention. 
We believe that our approach can be further extended to investigate beyond bilinguality in language models, providing valuable insights to the broader research community.


\section{Sparse Autoencoders}
\label{sec:sae}
A sparse autoencoder (SAE) is an autoencoder that enforces a sparsity constraint on its hidden layer.
In this study, we adopt a variant called TopK-SAE~\citep{makhzani2014ksparseautoencoders}, where the TopK activation function is applied at the hidden layer. 
Compared to a ReLU-based SAE~\citep{bricken2023monosemanticity, huben2024sparse}, TopK-SAE has been shown to be easier to train while maintaining sparsity and achieving higher reconstruction performance~\citep{gao2025scaling}.

Let $x \in \mathbb{R}^d$ be the input vector of an SAE and $n$ be the dimension of its hidden layer.
The encoder $E$ and decoder $D$ are defined as follows:
\begin{align}
    E(x) &= \mathrm{TopK}\bigl(W_{\mathrm{enc}}(x - b_{\mathrm{pre}})\bigr),\\
    \hat{x} &= D(E(x)) = W_{\mathrm{dec}} E(x) + b_{\mathrm{pre}},
\end{align}
where $W_{\mathrm{enc}} \in \mathbb{R}^{n \times d}$ and $W_{\mathrm{dec}} \in \mathbb{R}^{d \times n}$ are learned linear layers, and $b_{\mathrm{pre}} \in \mathbb{R}^d$ is a learnable bias parameter. 
$W_{\mathrm{dec}}$ is initialized as the transpose of $W_{\mathrm{enc}}$, and $b_{\mathrm{pre}}$ is initialized to the geometric median of the input data.

The training objective is the following mean squared error (MSE) loss:
\begin{align}
    L = \|x - \hat{x}\|^2_2. \label{eq:loss}
\end{align}

In this study, we control TopK-SAE by two hyperparameters:
$n$, the dimension of the hidden layer, 
and $K$, the number of hidden dimensions to keep active.
Interpreting $W_{\mathrm{dec}}$ as $n$ distinct vectors in $\mathbb{R}^d$, TopK-SAE can be seen as selecting $K$ vectors from $n$ and using their weighted sum to reconstruct the input. 
In this study, we denote each dimension of the encoder output $E(x) \in \mathbb{R}^n$ as a \emph{feature}.
We say the feature is \emph{activated} when it is selected during the TopK operation (i.e., utilized in reconstruction).

\section{Experiments}
\label{sec:exp}
In this section, we describe our experimental setup for analyzing the internal representations of bilingual language models using SAEs.
We detail the language models, datasets, and SAE training procedure (\cref{sec:exp_setup}); the procedure to find activation patterns of individual features (\cref{sec:find_patterns}), and the evaluation of language and concept selectivity of individual features (\cref{sec:language,sec:concept}).

\subsection{Experimental Setup}
\label{sec:exp_setup}

\paragraph{Language Models}
We used the models in the LLM-jp family (150M, 440M, 980M, 1.8B, 3.7B) as our focus for analysis~\citep{llmjp}.
These models were trained on the LLM-jp Corpus v3\footnote{\url{https://gitlab.llm-jp.nii.ac.jp/datasets/llm-jp-corpus-v3}}, which contains 1.7T tokens: 950B in English, 592B in Japanese, 114B in code, 0.8B in Korean, and 0.3B in Chinese.
We chose the LLM-jp family because (i) its intermediate checkpoints are (or available upon request) publicly available, (ii) it offers a range of model sizes, and (iii) it demonstrates bilingual capabilities in both English and Japanese.
We analyzed all of the layers of each language model.
For additional details about the models, please refer to the original repository\footnote{\url{https://huggingface.co/llm-jp/llm-jp-3-3.7b}}.

\paragraph{Datasets}
We train SAEs with the Japanese and English Wikipedia subsets in the LLM-jp Corpus v3.
For each document, we extract the first 64 tokens as the input to the language model, discard the \texttt{[BOS]} token representation, and apply L2 normalization to the remaining 63 representations ($\in\mathbb{R}^{63\times d}$), which serve as inputs to the SAE.
We use 100M tokens (50M in Japanese and 50M in English) for training, and 10M tokens (5M in Japanese and 5M in English) for evaluation.

\paragraph{TopK-SAE}
We use TopK-SAE and set the sparsity parameter $K=32$ and the hidden layer's dimension $n=32,768$ for all our experiments.
The batch size is fixed at 32,768, with a warm-up phase of 500 steps.
We perform a grid search to optimize the learning rate (\cref{app:lr}).
Training a single SAE takes approximately 10 minutes to 1.5 hours on a single A100 40GB GPU.
This variation is primarily due to the size of the Language Model (LM), as we simultaneously obtain intermediate activations through an LM while training an SAE.
Our implementation leverages the activation buffer to temporarily store a batch of LM activations, which are then used for SAE training~\citep{neelnandarep, marks2024dictionary-learning}.
The number of stored activations is adjusted according to the model size (see \cref{app:time_train_stock} for details).


\subsection{Finding Activation Patterns}
\label{sec:find_patterns}
We collect tokens that strongly activate each feature.
Specifically, we first determine the maximum activation value of each feature.
The threshold is then set at 70\% of this maximum value, and all tokens that exceed this threshold are collected from the evaluation set.

Next, we define token attribution distribution for feature $i$, denoted $f(v|i)$ for $1\leq i \leq n$, as the probability that an activation of feature $i$ was caused by token $v$. 
This is defined by the count of $v$ activating feature $i$ divided by the total number of feature $i$ being activated, satisfying $\sum_{v\in V}f(v|i) = 1$

We also assess the language distribution conditioned on the activation of each feature $i$. 
Specifically, we define $p(\text{en}|i)$ and $p(\text{ja}|i)$ as the probabilities that the input of the LM was in English or Japanese, respectively, given that the feature was activated, satisfying $p(\text{en}|i) + p(\text{ja}|i) = 1$.


\subsection{Language Selectivity Metrics}
\label{sec:language}
We classify each feature into three categories --- English Feature, Japanese Feature, and Mixed Feature --- based on the calculated language probability $p(\text{en}|i)$ and $p(\text{ja}|i)$.
The $i$-th feature is classified as an English Feature if $p(\text{en}|i) > 0.9$, a Japanese Feature if $p(\text{ja}|i) > 0.9$, and a Mixed Feature if neither condition is met. 
This classification reflects the dominant language context in which each feature is most strongly activated.

\subsection{Concept Selectivity Metrics}
\label{sec:concept}
\begin{figure}[tbp]
    \centering
    \includegraphics[width=\columnwidth]{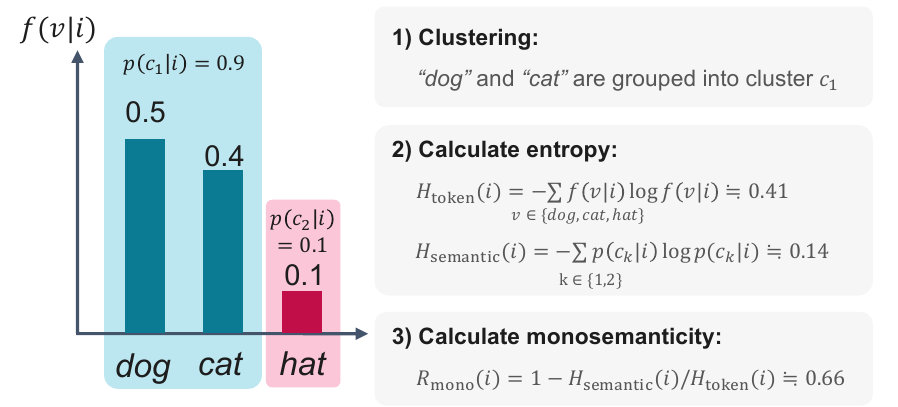}
    \caption{The procedure for calculating Monosemanticity ($R_{\mathrm{mono}}(i)$) from  Token Entropy ($H_{\mathrm{token}}(i)$) and Semantic Entropy ($H_{\mathrm{semantic}}(i)$) for the $i$-th feature.}
    \label{fig:mono}
\end{figure}

\begin{figure*}[tbp]
    \centering
    \includegraphics[width=\textwidth]{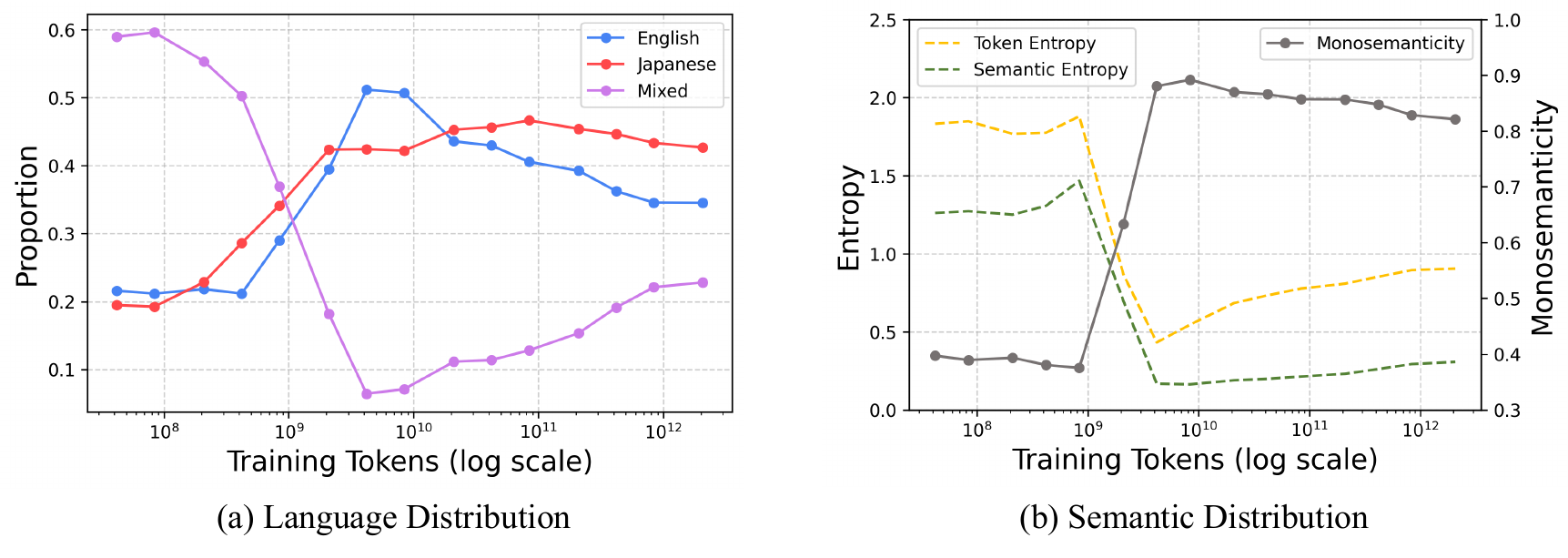}
    \caption{(a) Language Distribution and (b) Semantic Distribution of SAE's features at the 14th layer of the 3.7B model across training stages.
    During early training ($\leq 4 \times 10^8$ tokens), the model exhibits a high proportion of mixed language features and low monosemanticity, indicating that features are activated by tokens from both languages without clear semantic coherence.
    As training continues ($4 \times 10^8$ -- $4 \times10^9$ tokens), the mixed language proportion decreases while monosemanticity increases, reflecting more language-specific and semantically coherent features.
    In the late training stage ($\geq 4 \times 10^9$ tokens), the mixed-language proportion rises again, but high monosemanticity is maintained, suggesting the emergence of bilingual semantic representations.
    }
    \label{fig:lang_sem_3.7b}
\end{figure*}
To quantitatively evaluate the semantic alignment of feature-activating tokens (i.e., tokens that activate a certain feature) over languages, we use three metrics: Token Entropy, Semantic Entropy, and Monosemanticity.

\paragraph{Token Entropy}
Token Entropy measures the diversity of tokens that activate a given feature.
For the $i$-th feature, it is calculated as:
\begin{align}
    H_{\text{token}}(i) = - \sum_{v \in V} f(v|i) \log f(v|i)
\end{align}
A high Token Entropy $H_{\text{token}}(i)$ value indicates that a wide variety of tokens can activate the feature, while a low value suggests that only a limited set of tokens do so.

\paragraph{Semantic Entropy}
Semantic Entropy quantifies the diversity of semantic meanings among the tokens that activate each feature.
Calculating Semantic Entropy consists of three steps: embedding tokens, clustering based on cosine similarity, and computing the entropy of the resulting clusters.

\begin{enumerate}
    \item \textbf{Token Embedding}:
    Token embeddings of feature-activating tokens, or tokens that activated feature $i$ at least once, are extracted from the embedding layer of the 3.7B model.

    \item \textbf{Semantic Clustering}:
    Using the extracted embeddings, tokens with a cosine similarity above a predefined threshold are grouped into the same semantic cluster\footnote{We set the cosine similarity threshold at 0.1 because it effectively balances capturing semantically related tokens and avoiding over-clustering of unrelated tokens.}.

    \item \textbf{Entropy Calculation}:
    Similar to Token Entropy, we compute the entropy over these semantic clusters using the formula:
    \begin{equation}
        H_{\text{semantic}}(i) = - \sum_{c \in C_i} p(c|i) \log p(c|i)
    \end{equation}
    where \( C_i \) is the set of semantic clusters for the \( i \)-th feature, and \( p(c|i) \) is the probability that an activation of feature $i$ was caused by a token belonging to cluster \( c \).
\end{enumerate}

A high value of \( H_{\text{semantic}}(i) \) indicates that the activating tokens are semantically diverse, while a low value suggests they are semantically consistent.
For example, in \cref{fig:mono}, ``dog'' and ``cat'' are grouped into the same cluster, resulting in a relatively low semantic entropy of $H_{\mathrm{semantic}} = 0.14$.
This entropy effectively captures the degree of semantic diversity in token activation patterns.

This quantification is based on the approach proposed by \citet{farquhar_detecting_2024}.
While they used semantic entropy to assess the semantic diversity among sentences and leveraged LLMs to cluster these sentences, our method applies semantic entropy to measure semantic diversity among tokens.

\paragraph{Monosemanticity}

Monosemanticity provides a normalized measure that quantifies the relationship between the semantic diversity and token diversity.
It is defined as the complement of the ratio of semantic entropy to token entropy:
\begin{align}
R_{\text{mono}}(i) = 1- \frac{H_{\text{semantic}}(i)}{H_{\text{token}}(i)}
\end{align}
This ratio ranges between 0 and 1:
A value close to 1 suggests that although the feature is activated by a wide variety of tokens (high Token Entropy), these tokens are semantically similar (low Semantic Entropy).
A value close to 0 indicates that the activating tokens are both diverse in form and meaning (high token entropy and high semantic entropy) or they are both consistent in form and meaning (low token entropy and low semantic entropy).
In the special case where $H_{\text{token}}(i) = 0$ (i.e., only one token activates the feature), we define $R_{\text{mono}}(i) = 1$.


\section{Observations}
\label{sec:results}
We first examine the internal representations of the model by analyzing the distribution of each SAE's features trained on various checkpoints (\cref{sec:ckpt}), layers (\cref{sec:layers}), and model sizes (\cref{sec:sizes}).

\subsection{LLMs first learn languages independently before aligning them bilingually}
\label{sec:ckpt}

\begin{figure}[tbp]
\centering
\begin{overpic}[width=\columnwidth]{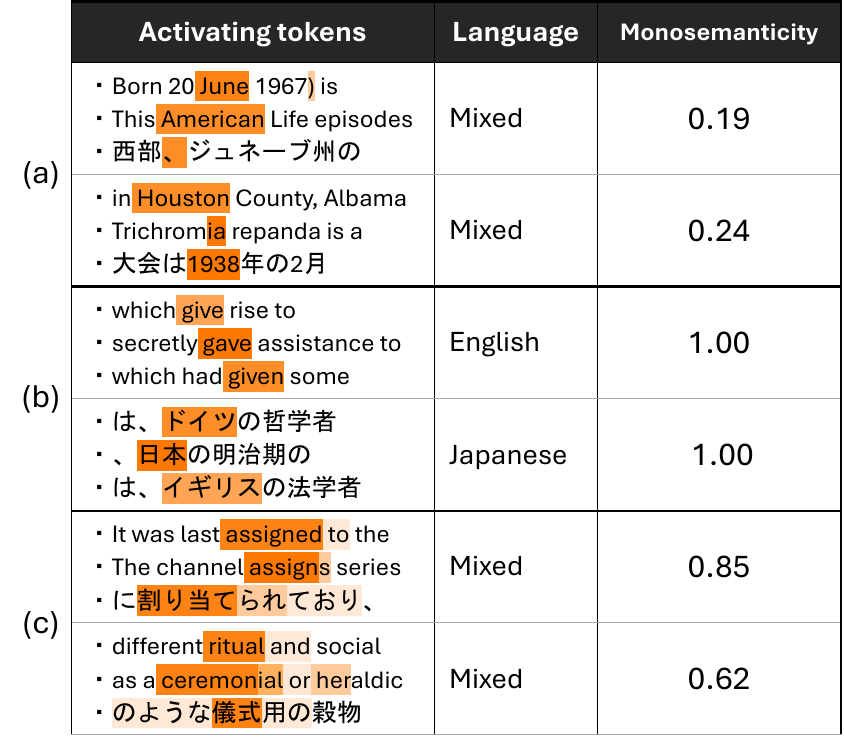}
    \put(0,90){\hypertarget{fig:ex-a}{}}
    \put(0,90){\hypertarget{fig:ex-b}{}}
    \put(0,90){\hypertarget{fig:ex-c}{}}
\end{overpic}
\caption{
Activation patterns of features at the 14th layer of the 3.7B model across training stages.
(a) In the early training stage ($4 \times 10^6$ tokens), features are activated by random tokens without any clear semantic structure.
(b) In the mid-training stage ($4 \times 10^9$ tokens), features become more language-specific, with tokens activating on semantically similar words in a single language.
(c) In the fully trained model ($2 \times 10^{12}$ tokens), features exhibit bilingual activation, with semantically related tokens appearing in both Japanese and English.
}
\label{fig:ex}
\end{figure}

\cref{fig:lang_sem_3.7b} presents the evolution of language and semantic distributions for SAE's features at the 14th layer of the 3.7B model across different training stages. 
In the early training phase ($\leq4 \times 10^8$ tokens), most features are categorized as mixed features and exhibit low monosemanticity.
This indicates that individual features are activated by tokens from both Japanese and English without any consistent semantic pattern, effectively behaving as random activation patterns.
This observation is consistent with the activation patterns shown in \hyperlink{fig:ex-a}{Figure~\ref*{fig:ex}(a)}, where activated tokens lack any clear semantic or linguistic coherence.

As training progresses into the middle phase ($4 \times 10^8$ -- $4 \times 10^9$ tokens), the proportion of mixed language features sharply declines, while monosemanticity markedly increases. 
This shift suggests that features become more language-specific, activating on tokens within a single language that share coherent semantic meanings.
For instance, \hyperlink{fig:ex-b}{Figure~\ref*{fig:ex}(b)} illustrates two representative examples: the first feature is activated by English tokens ``give,'' ``gave,'' and ``given,'' which are grammatical variations of the same verb, while the second feature is activated by Japanese tokens representing country names (``ドイツ'' for Germany, ``日本'' for Japan, and ``イギリス'' for the United Kingdom). 
These patterns demonstrate that the model is beginning to organize and align semantics within each language independently.

In the late training stage ($\geq 4 \times 10^9$ tokens), the model exhibits a resurgence of mixed-language features while maintaining high monosemanticity.
This phase signifies a transition from language-specific semantics to bilingual semantic alignment, where features activate on semantically similar tokens across both languages.
As shown in \hyperlink{fig:ex-c}{Figure~\ref*{fig:ex}(c)}, one feature is activated by ``assigned,'' ``assign,'' and ``割り当て'' (the Japanese term for ``assign''), while another is activated by ``ritual,'' ``ceremon,'' and ``儀式'' (the Japanese term for ``ritual'').
These examples confirm that the model now captures semantic correspondences between languages, functioning as a bilingual representation.

These findings suggest that LLMs learn in two distinct stages. 
\begin{enumerate}
    \item During the early to mid-training phase, they develop independent semantic representations within each language.
    \item In the subsequent mid-to-late training phase, they begin to align these semantic representations across languages
\end{enumerate}


\subsection{Mid-layers capture more bilingual alignments}
\label{sec:layers}

\begin{figure}[tbp]
    \centering
    \includegraphics[width=\columnwidth]{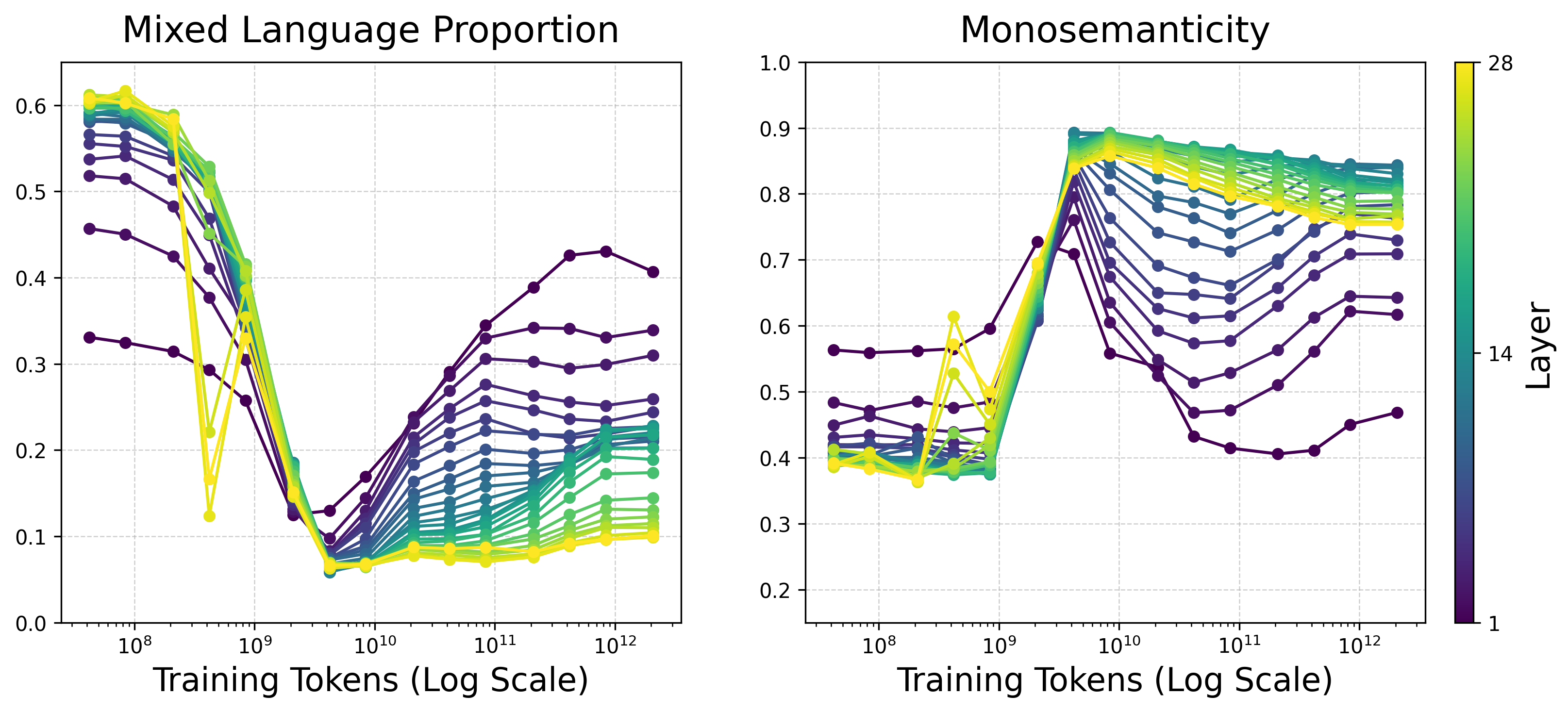}
    \caption{Layer-wise evolution of mixed language proportion and the monosemanticity in 3.7B model across training stages.}
    \label{fig:mix_all_3.7b}
\end{figure}

\begin{figure}[tbp]
    \centering
    \includegraphics[width=\columnwidth]{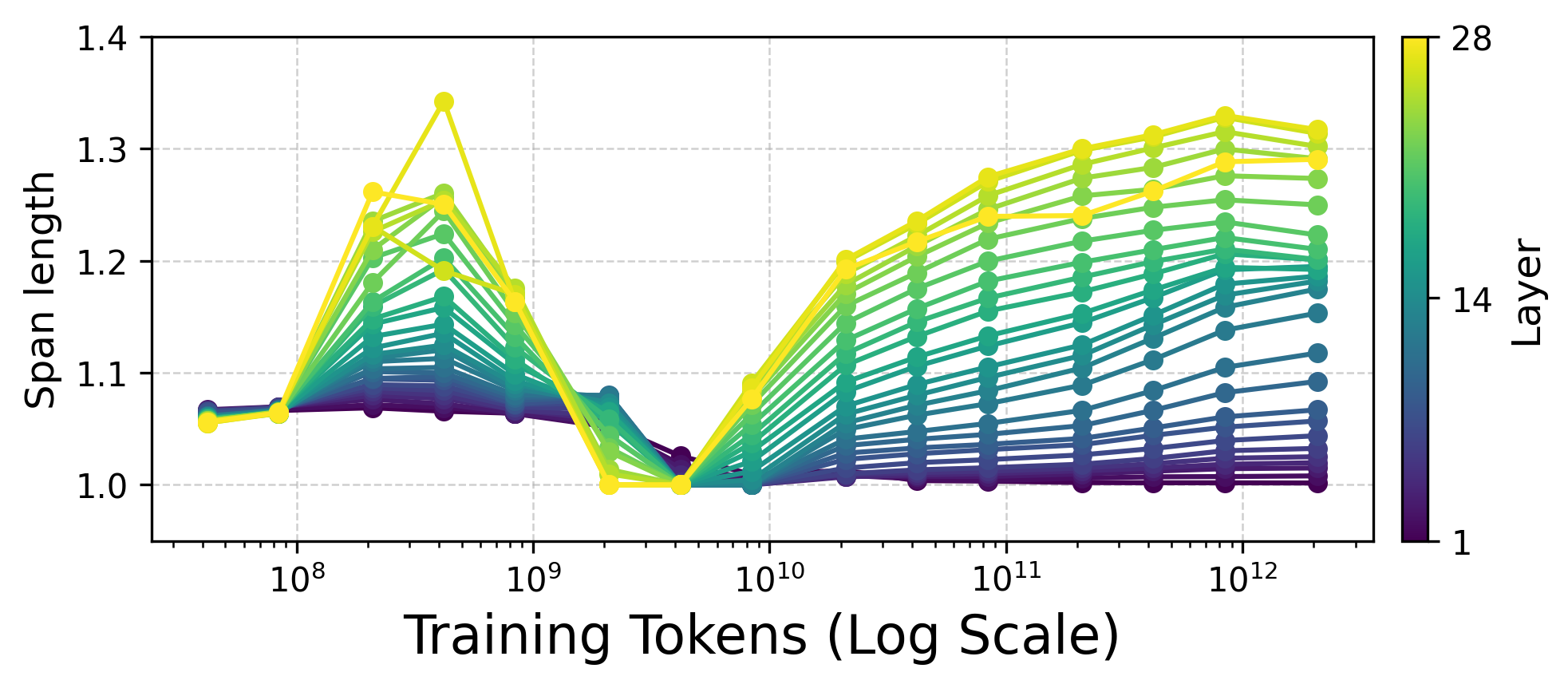}
    \caption{Layer-wise evolution of span length average in 3.7B model across training stages.}
    \label{fig:span_avg_3.7b}
\end{figure}

\cref{fig:mix_all_3.7b} illustrates the layer-wise evolution of the mixed language proportion and the monosemanticity of SAEs' features in the 3.7B model across training stages.
In the early to mid-training phase ($\leq 4 \times 10^9$ tokens), all layers exhibit a decrease in the mixed language proportion and an increase in monosemanticity. 
This suggests that the model initially learns the semantics within each language in all layers.

As training progresses into the later stages, layer behaviors begin to diverge. 
The mid layers (green) align with the behavior of the 14th layer described in \cref{sec:ckpt}, while the lower (purple) and upper layers (yellow) follow distinct patterns.

\begin{figure}[tbp]
    \centering
    \begin{overpic}[width=\columnwidth]{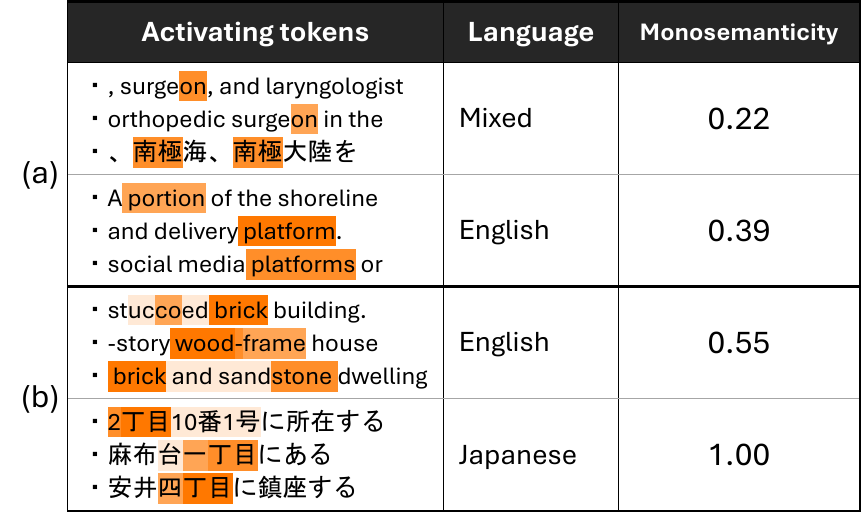}
        \put(0,60){\hypertarget{fig:ex-lowup-a}{}}
        \put(0,60){\hypertarget{fig:ex-lowup-b}{}}
    \end{overpic}
    \caption{Activation patterns of features in the 3.7B model.
    (a) In the lower layer (2nd layer), features exhibit activation across multiple meanings.
    (b) In the upper layer (26th layer), features primarily activate on long-span tokens.
    }
    \label{fig:ex_lowup}
\end{figure}

In the lower layers, particularly the initial layers, the mixed language proportion increases, while monosemanticity decreases compared to the mid layers.
This suggests a tendency toward polysemanticity, where a single feature is activated by multiple meanings. 
As illustrated in \hyperlink{fig:ex-lowup-a}{Figure ~\ref*{fig:ex_lowup}(a)}, the first feature is activated on both the English word ``on'' and ``南極'' (Japanese for ``Antarctica''), and the second feature is activated on `` portion'', `` platform'', and `` platforms''.
Although these activation patterns are less random than in the early training stages, they still occur across multiple tokens, reflecting the model's polysemantic nature in these layers.
Such behavior can be attributed to the model's proximity to the input layer, where it must distinguish between a vast vocabulary of approximately 100,000 tokens, which exceeds the dimension $n$ of the intermediate layers of the SAE.

On the other hand, the upper layers consistently maintain a lower mixed language proportion than the mid layers, while their monosemanticity declines even further as training progresses.
Analyzing span length --- the number of consecutive tokens each feature activates --- reveals that these deeper layers increasingly focus on longer spans (\cref{fig:span_avg_3.7b}), indicating that features are not monosemantic at the token level because they span multiple, contextually connected tokens.
For instance, as shown in \hyperlink{fig:ex-lowup-b}{Figure~\ref*{fig:ex_lowup}(b)}, the first feature is activated on phrases such as ``uccoed brick'', ``wood-frame'', and ``brick and sandstone'', all referring to building materials with spans of around three tokens.
The second feature activates on Japanese addresses such as ``2丁目10番1号'' (similar to ``Block 2, No. 10-1''), ``一丁目'' (``Block 1''), ``四丁目'' (``Block 4''), each spanning multiple tokens.

\begin{figure}[tbp]
    \centering
    \includegraphics[width=\columnwidth]{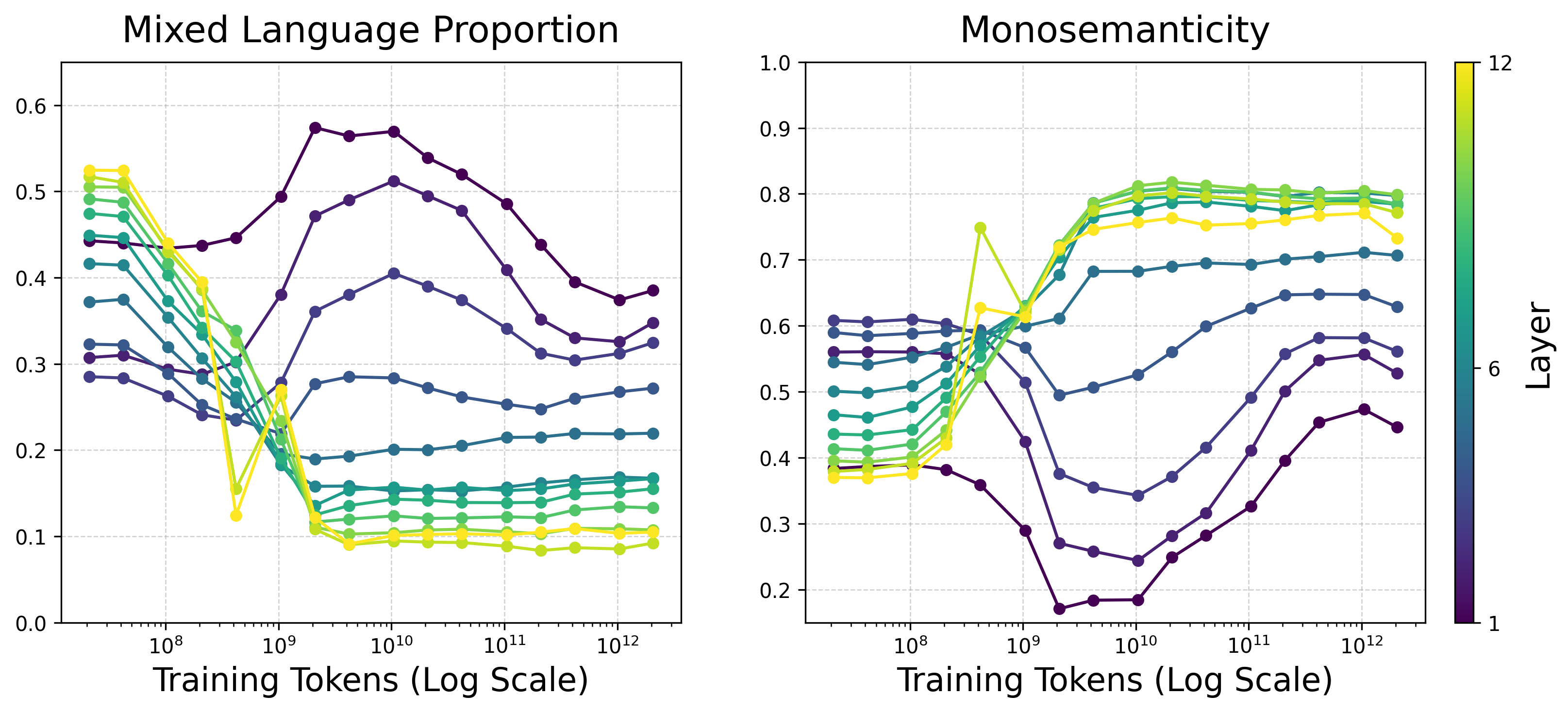}
    \caption{Layer-wise evolution of the mixed language proportion and the monosemanticity in the 150M model across training stages.}
    \label{fig:mix_mono_150m}
\end{figure}

\begin{figure}[tbp]
    \centering
    \includegraphics[width=\columnwidth]{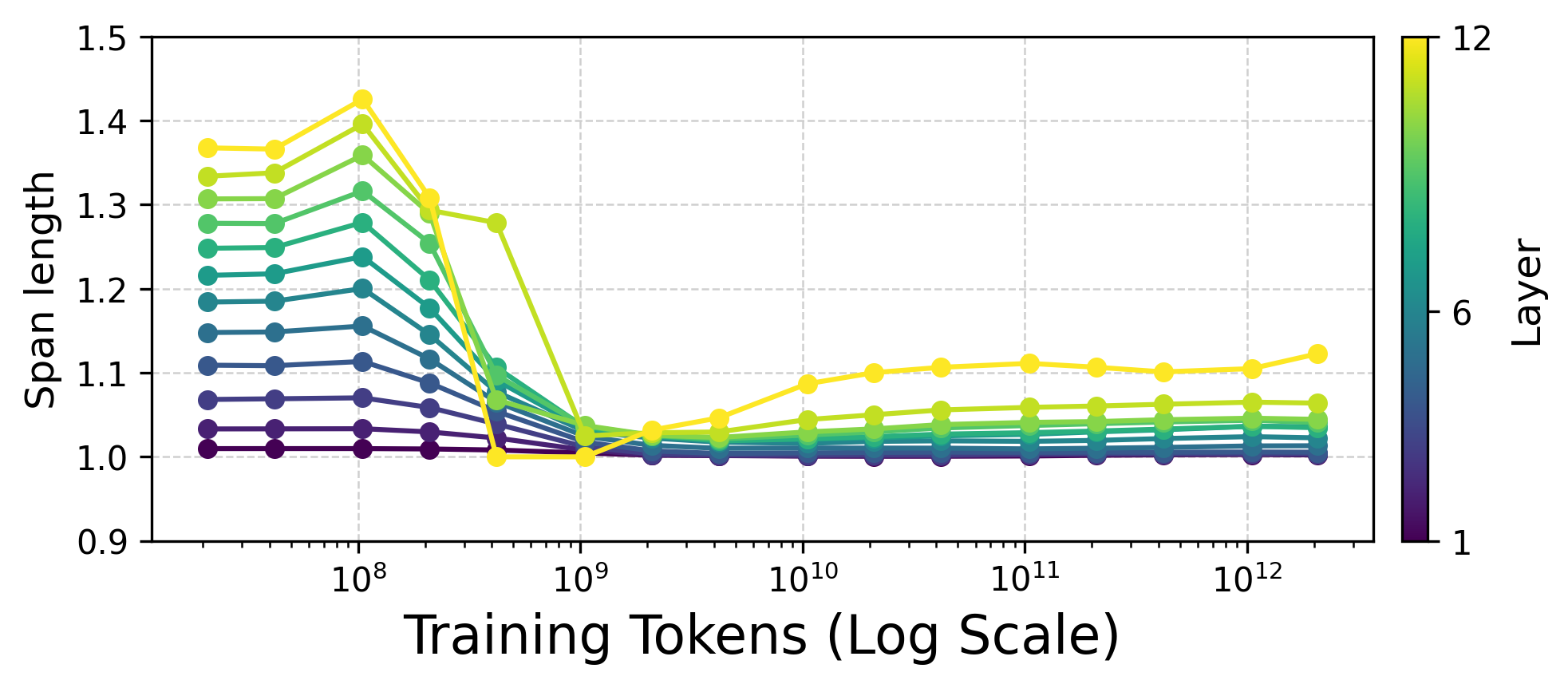}
    \caption{Layer-wise evolution of span length average in 150M model across training stages.}
    \label{fig:span_avg_150m}
    \vspace{-10pt}
\end{figure}

From these findings, it can be inferred that
\begin{itemize}
    \item Mid layers specialize in learning bilingual representations, balancing monosemanticity and mixed language proportion.
    \item Lower layers exhibit polysemanticity, distinguishing a wide variety of tokens in the vocabulary.
    \item Upper layers focus on multi-token concepts by capturing longer spans rather than individual tokens.
\end{itemize}

\subsection{Larger LMs develop more bilingual alignments}
\label{sec:sizes}

\cref{fig:mix_mono_150m} illustrates the layer-wise evolution of the mixed language proportion and the monosemanticity in the 150M model.
\cref{fig:mix_mono_440m,fig:mix_mono_980m,fig:mix_mono_1.8b} also show the result of other sizes (440M, 980M, and 1.8B).
In the early to mid-training phase ($ \leq 4 \times 10^9$ tokens), the behavior of around mid layers mirrors that of the 3.7B model: the mixed language proportion decreases while monosemanticity increases.
This indicates that even in smaller models, the early training stage primarily involves learning languages individually.

However, a divergence becomes apparent in three aspects: (1) within mid layers during the late training phase ($\geq 4 \times 10^9$ tokens), (2) within upper layers during the late training phase, and (3) within the lower layers during all training phases.

In the mid layers, the smaller model shows a smaller increase in mixed language proportion compared to the larger model, as described in Section \ref{sec:layers}.
The features learned by the smaller model around the mid layers are less inclined to exhibit high monosemanticity across languages.
This suggests that a much lower capacity for learning bilingual features compared to the larger model.

In the upper layers, monosemanticity is smaller compared to larger models at the late training stages. 
As shown in \cref{fig:span_avg_150m}, the layer-wise change in span length in the 150M model indicates that the increase in span length in the upper layers is also smaller than in larger models. 
These observations suggest that the upper layers in smaller models cannot capture the context-level meanings.

In lower layers, the smaller model retains a relatively high mixed language proportion and low monosemanticity. 
This indicates a failure to adequately capture semantics even within individual languages, unlike the larger model, where lower layers effectively acquire intra-language semantics.

In summary, two key observations can be drawn: 
\begin{itemize}
    \item Larger models exhibit a greater ability to learn bilingual features in the mid layers, while smaller models struggle to do so. 
    \item  Although smaller models may acquire some degree of semantic alignments within individual languages in certain layers, they lack a strong tendency to generalize these features towards bilingual representations in the later stages of training.
\end{itemize}

\section{Intervention}
\label{sec:add_bi}

\begin{figure}[tbp]
    \centering
    \includegraphics[width=\columnwidth]{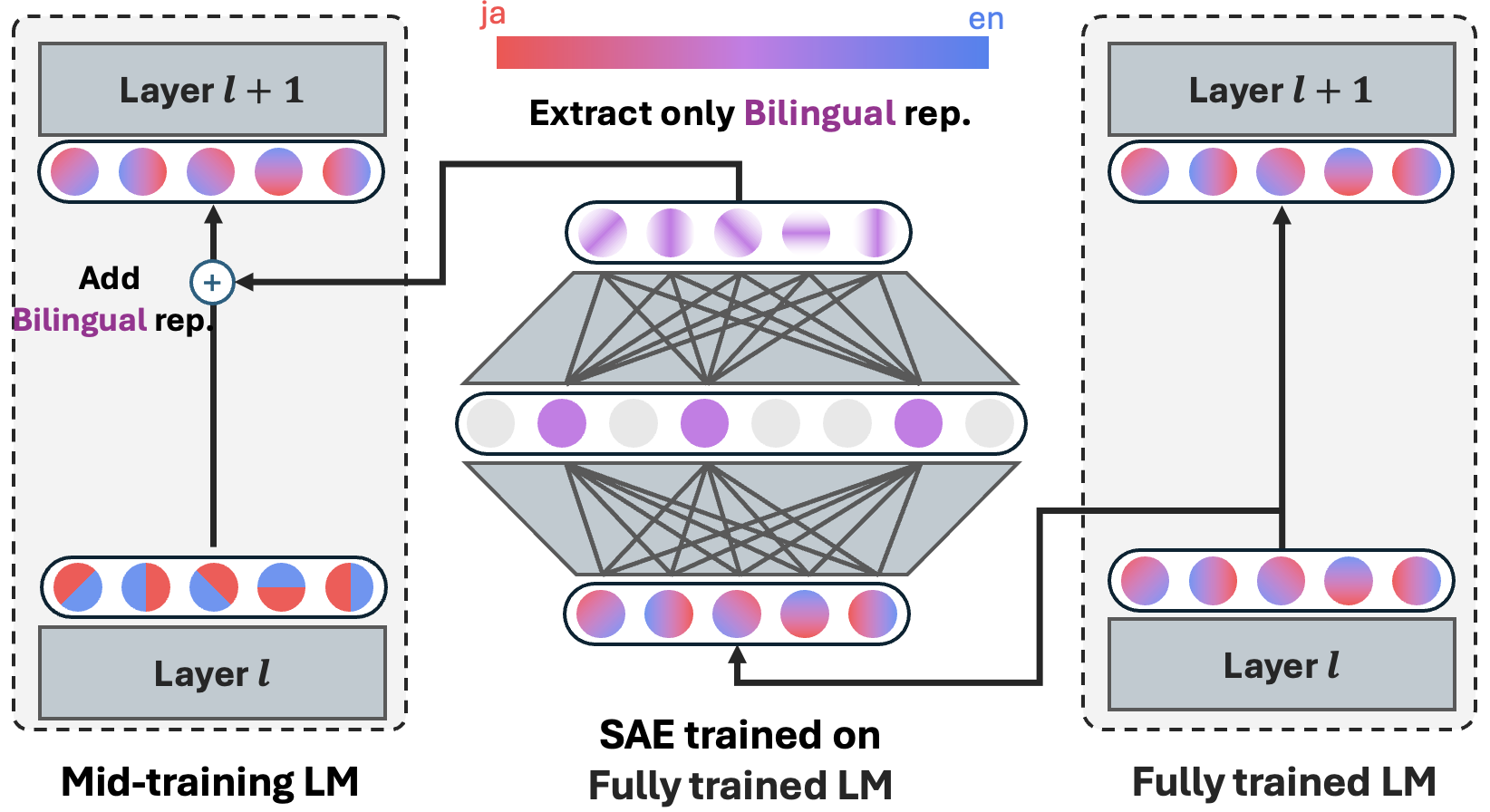}
    \caption{Illustration of adding bilingual representations from a fully trained model into a mid-training model.}
    \label{fig:add_bi}
    \vspace{-15pt}
\end{figure}

We hypothesize that bilingual representations, which correspond to bilingual features, play a crucial role in the performance of a fully trained model.
If this is true, integrating these representations into a mid-training model should significantly enhance its performance.
To test this, here, we extract bilingual representations from a fully trained model using a TopK-SAE and inject them into the intermediate representations of a mid-training model. 
This process is illustrated in \cref{fig:add_bi}.

\subsection{Method}
Mathematically, let $\bm{X}_{\text{full}}^\ell, \bm{X}_\text{mid}^\ell \in \mathbb{R}^{T \times d}$ denote the outputs of the $\ell$-th layer of the fully trained and mid-training models, respectively, where $T$ is the sequence length and $d$ is the model dimension.
We also denote $E: \mathbb{R}^d \rightarrow \mathbb{R}^n$ and $D: \mathbb{R}^n \rightarrow \mathbb{R}^d$ as an encoder and a decoder of TopK-SAE trained on the fully trained model.
A binary mask $\mathbf{mask} \in \mathbb{R}^n$ is also defined, with $m$ elements set to 1 and others to 0, forcing only the bilingual features to get activated.
The intervention is formulated as follows:
\begin{align}
    \bm{X}_\text{mid}^\ell \leftarrow \bm{X}_\text{mid}^\ell + \alpha \cdot D (\mathbf{mask} \odot E(\bm{X}_{\text{full}}^\ell))
\end{align}
where $\alpha$ is a hyperparameter controlling the strength of the intervention, set to 0.1 in our experiments (see \cref{app:alpha_m} for the result of other values).
This method allows us to assess the direct impact of the bilingual representation incorporation.

\subsection{Setup}
We conducted experiments using the 14th layer of the 3.7B model.
As the mid-training model, we selected the checkpoint at 10,000 (approximately 40B training tokens), where the mixed language proportion in this layer is relatively low (\cref{fig:lang_sem_3.7b}).
We evaluated the effects of three feature types: English, Japanese, and Bilingual (Mixed).
The number of selected feature dimensions was set to $m=5,000$ (see \cref{app:alpha_m} for the result of other values ).
Each setting was evaluated five times, and the results were averaged.

\subsection{Results \& Discussion}
\begin{table}[tbp]
\centering
\small
\begin{tabular}{crrrrr} 
\toprule
& \multicolumn{3}{c}{Perplexity ($\downarrow$)} & \multicolumn{2}{c}{COMET-22 ($\uparrow$)} \\
\cmidrule(lr){2-4} \cmidrule(lr){5-6}
Add & En & Ja & all & En $\to$ Ja & Ja $\to$ En \\
\cmidrule(lr){1-6}
- & 18.70 & 25.78 & 22.43 & 61.1 & 56.4 \\
\cmidrule(lr){1-6}
En & \gcella18.53 & 25.64 & \gcella22.28 & \gcella61.4 & \gcella56.9 \\
Ja & 18.65 & \gcellc25.33 & \gcellb22.17 & \gcella61.3 & \gcellb\textbf{57.2} \\
Bi & \gcellb\textbf{18.36} & \gcellc\textbf{25.20} & \gcellb\textbf{21.96} & \gcellc\textbf{62.5} & \gcellb\textbf{57.2} \\

\bottomrule
\end{tabular}
\caption{Baseline denotes the perplexity (PPL) of the mid-training model without any intervention. 
Adding mixed (bilingual) representations leads to a greater reduction in PPL compared to adding Japanese or English representations.}
\vspace{-15pt}
\label{tab:add_bi}
\end{table}

\cref{tab:add_bi} shows the results.
Adding English-specific representations mainly improved English performance, while adding Japanese-specific representations primarily enhanced Japanese performance. 
In contrast, adding bilingual representations significantly improved both languages' performances.
This performance boost holds even when varying the hyperparameters $\alpha$ and $m$ as shown in \cref{tab:alpha_m}.
These results support our hypothesis that the bilingual alignments acquired by the model in the later training stages play a crucial role in its performance.

Note that this method requires the output of Layer $\ell$ from the fully trained model, meaning that the SAE alone cannot directly enhance the performance of a mid-training model.
However, our findings reveal that the bilingual information encoded in the later training stages is more critical for performance than monolingual information. 
This suggests that designing a training schedule that encourages the acquisition of bilingual knowledge in the later stages of pre-training could be beneficial.


\section{Related Work}
\label{sec:rel_work}
Understanding the internal mechanisms of LLMs has become a major focus of the research community.
Recent studies show neural networks can represent more features than their dimensions~\citep{elhage2022superposition}.
To disentangle these representations, SAEs have emerged as a key tool for decomposing them into interpretable components~\citep{huben2024sparse, Olshausen1997Sparse}.
While early work primarily focused on a single SAE, recent studies have shifted toward comparing SAE features across layers~\citep{balcells2024evolutionsaefeatureslayers, balagansky2025mechanistic}, model architectures~\citep{lan2024sparseautoencodersrevealuniversal, lindsey2024crosscoders}, or fine-tuning stages~\citep{lindsey2024crosscoders, wang2025towards}.
\citet{xu2024trackingfeaturedynamicsllm} concurrently tracks feature formation during training, but lacks quantitative evaluation. 

Another line of research has explored the multilingual capability of language models. 
\citet{zeng-etal-2025-converging} explored the formation of multilingual capabilities through neuron-level analysis and showed that as models become larger and training progresses, they exhibit an increasing degree of multilingual understanding.
This result aligns with our SAE-based analysis results.
\citet{wang2024sharing} identified neurons shared across languages and tasks, while \citet{tang-etal-2024-language} and \citet{kojima-etal-2024-multilingual} highlighted language-specific neurons, demonstrating their impact on model performance and language output.

Our research builds on these foundations and contributes to them in three key ways: (1) we investigate the formation process of bilingual capabilities within a bilingual language model, (2) we conduct a comparative analysis across training stages, model sizes, and layers, and (3) we exmploy SAEs to perform direct interventions on bilingual representations, offering novel insights on the dynamics of bilingual representation in language models.


\section{Conclusion}
\label{sec:conclusion}
In this study, we investigated the evolution of internal representations in language models using SAEs.
Our analysis revealed that bilingual language models initially learn languages independently and later develop bilingual alignments, particularly in the mid-layers of larger models.
We further demonstrated the importance of bilingual representations by conducting targeted interventions with SAEs.
Beyond using SAEs solely for interpreting language models, we leveraged them to manipulate internal representations, showcasing their potential as a tool for both analysis and intervention.
We believe that our approach can be extended to explore beyond analyzing the bilinguality of language models and offer valuable insights for the broader research community.

\section{Limitations}
This study explored the internal mechanisms of bilingual language models, specifically focusing on English, Japanese, and their bilingual interactions. 
While this provides insights into cross-lingual representation between these two typologically distinct languages, the findings may not generalize to all language pairs.
Future research should investigate a wider range of language pairs to validate and extend our observations.

Another limitation is the interpretability of the SAEs used in our analysis.
While SAEs allowed us to investigate the types of information that models tend to encode as features, recent studies have raised concerns about the reliability and interpretability of them.
Additionally, given that the reconstruction accuracy was not perfect, our analysis is based on an approximation of the model's internal representations.
As a direction for future work, combining SAEs with other analytical methods could lead to a more robust and comprehensive understanding of the model's behavior.

\section*{Acknowledgements}
This work was supported by JSPS KAKENHI (Grant Number JP25KJ06300) and JST BOOST, Japan (Grant Number JPMJBS2412). 
We used the ``mdx: a platform for building a data-empowered society''~\citep{mdx}. 
ChatGPT was employed to assist with manuscript preparation and code generation, and all outputs were thoroughly reviewed and validated by the authors.

\todo{77行目のversionwithcommentsをコメントアウトする}
\ifdefined\VersionWithComments%
\listoftodos    
\else
\fi


\bibliography{custom}
\appendix
\clearpage

\section{Training Details}
\label{app:train_detail}
\subsection{Learning Rate Selection}
\label{app:lr}

We determined the optimal learning rate for training SAEs on each LM size by a grid search.
Specifically, we tested several learning rates (1e-4, 2e-4, 5e-4, 1e-3, 2e-3, 5e-3) for each LM, the last checkpoint, and the middle layer (maxlayer // 2), and selected one that resulted in the lowest reconstruction loss (\cref{eq:loss}) on the validation set.

\begin{figure}[htbp]
    \centering
    \includegraphics[width=\columnwidth]{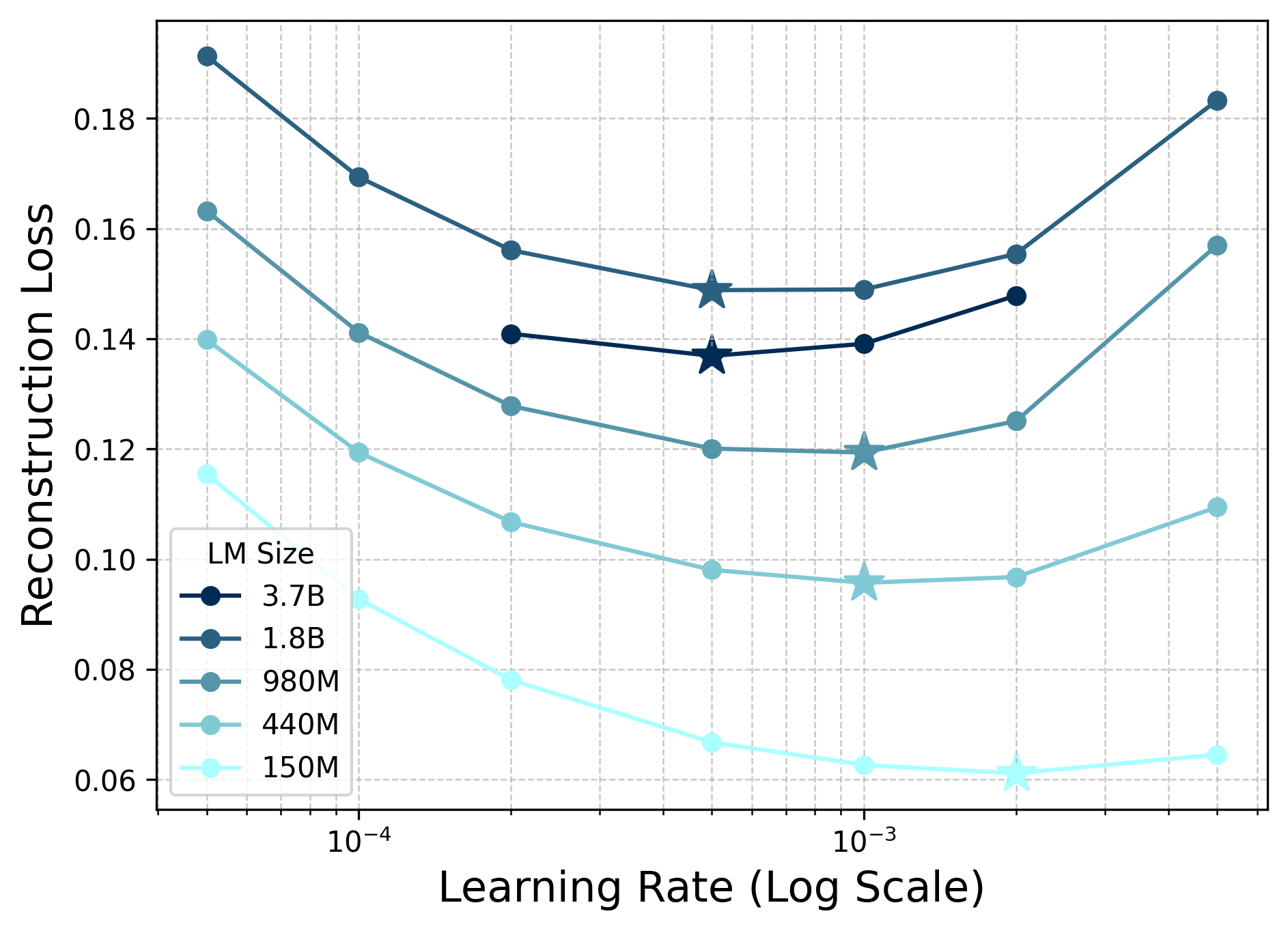}
    \caption{Learning Rate vs. Reconstruction Loss for SAEs on Various Model Sizes. The star markers indicate the lowest loss points for each model.}
    \label{fig:lr}
\end{figure}

~\cref{fig:lr} shows the result.
Our experiments revealed that smaller learning rates were more effective for training SAEs on larger LMs.
The selected learning rates for each model size are summarized in ~\cref{tab:lr}.

\begin{table}[htbp]
\centering
\begin{tabular}{lc}
\hline
LM Size & Optimal Learning Rate \\ 
\hline
150M & 2e-3 \\
440M & 1e-3 \\
980M & 1e-3 \\
1.8B & 5e-4 \\
3.7B & 5e-4 \\
\hline
\end{tabular}
\caption{Optimal learning rates for training SAEs across different LM sizes}
\label{tab:lr}
\end{table}

\subsection{Time for training SAEs \& the number of stored activations}
\label{app:time_train_stock}
\cref{tab:time_stock} shows the details.

\begin{table}[htbp]
\centering
\begin{tabular}{lcc}
\hline
LM Size & Training Time & N of act. \\ 
\hline
150M & 20min & 10M \\
440M & 25min & 5M \\
980M & 40min & 2M \\
1.8B & 60min & 1M \\
3.7B & 90min & 0.5M \\
\hline
\end{tabular}
\caption{The training time for each SAE and the number of buffered activations for each model size.}
\label{tab:time_stock}
\end{table}

\section{Ablation Study of Adding Bilingual Features}
\label{app:alpha_m}
\cref{tab:alpha_m} shows the ablation result of different $alpha$ and $m$.

\begin{table}[htbp]
\centering
\begin{tabular}{cccrrr} 
\toprule
&&& \multicolumn{3}{c}{Perplexity (dif.)} \\
\cmidrule(lr){4-6}
$\alpha$ & $m$ & Add & En & Ja & all \\
\cmidrule(lr){1-6}
&& - & $17.57$ & $19.54$ & $15.39$ \\
\cmidrule(lr){1-6}
\multirow{9}{*}{0.05} & \multirow{3}{*}{1000} & En & $-0.08$ & $-0.06$ & $-0.07$ \\
&  & Ja & $-0.07$ & $-0.09$ & $-0.08$ \\
&  & Bi & $-0.10$ & $-0.10$ & $-0.10$ \\
\cmidrule(lr){2-6}
& \multirow{3}{*}{3000} & En & $-0.10$ & $-0.07$ & $-0.09$ \\
&  & Ja & $-0.08$ & $-0.15$ & $-0.12$ \\
&  & Bi & $-0.16$ & $-0.19$ & $-0.18$ \\
\cmidrule(lr){2-6}
& \multirow{3}{*}{5000} & En & $-0.12$ & $-0.08$ & $-0.10$ \\
&  & Ja & $-0.09$ & $-0.21$ & $-0.15$ \\
&  & Bi & $-0.21$ & $-0.28$ & $-0.25$ \\

\cmidrule(lr){1-6}

\multirow{9}{*}{0.10} & \multirow{3}{*}{1000} & En & $-0.08$ & $-0.07$ & $-0.08$ \\
&  & Ja & $-0.06$ & $-0.12$ & $-0.09$ \\
&  & Bi & $-0.12$ & $-0.15$ & $-0.14$ \\
\cmidrule(lr){2-6}
& \multirow{3}{*}{3000} & En & $-0.13$ & $-0.09$ & $-0.11$ \\
&  & Ja & $-0.08$ & $-0.25$ & $-0.17$ \\
&  & Bi & $-0.23$ & $-0.34$ & $-0.29$ \\
\cmidrule(lr){2-6}
& \multirow{3}{*}{5000} & En & $-0.16$ & $-0.11$ & $-0.14$ \\
&  & Ja & $-0.10$ & $-0.36$ & $-0.24$ \\
&  & Bi & $-0.33$ & $-0.50$ & $-0.42$ \\
    
\cmidrule(lr){1-6}

\multirow{9}{*}{0.20}& \multirow{3}{*}{1000} & En & $+0.13$ & $+0.14$ & $+0.13$ \\
&  & Ja & $+0.18$ & $+0.03$ & $+0.10$ \\
&  & Bi & $+0.05$ & $-0.03$ & $+0.01$ \\
\cmidrule(lr){2-6}
& \multirow{3}{*}{3000} & En & $+0.01$ & $+0.10$ & $+0.06$ \\
&  & Ja & $+0.15$ & $-0.25$ & $-0.06$ \\
&  & Bi & $-0.18$ & $-0.40$ & $-0.30$ \\
\cmidrule(lr){2-6}
& \multirow{3}{*}{5000} & En & $-0.07$ & $+0.05$ & $-0.01$ \\
&  & Ja & $+0.10$ & $-0.49$ & $-0.21$ \\
&  & Bi & $-0.37$ & $-0.72$ & $-0.56$ \\

\bottomrule
\end{tabular}
\caption{Baseline denotes the perplexity (PPL) of the mid-training model without any intervention.}
\label{tab:alpha_m}
\end{table}

\newpage

\begin{figure*}[tbp]
    \centering
    \includegraphics[width=\textwidth]{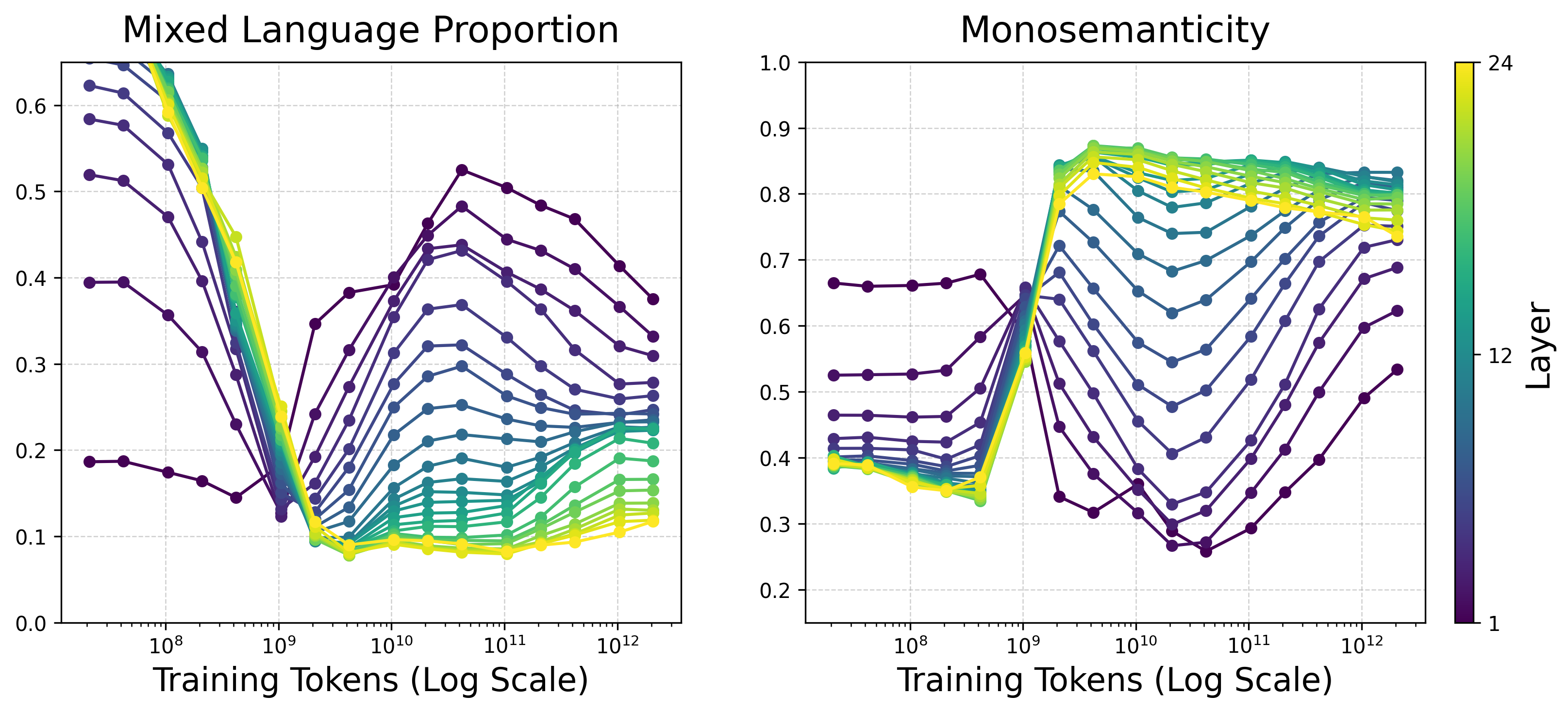}
    \caption{Layer-wise evolution of the mixed language proportion and the monosemanticity in the 1.8B model across training stages.}
    \label{fig:mix_mono_1.8b}
\end{figure*}

\begin{figure*}[tbp]
    \centering
    \includegraphics[width=\textwidth]{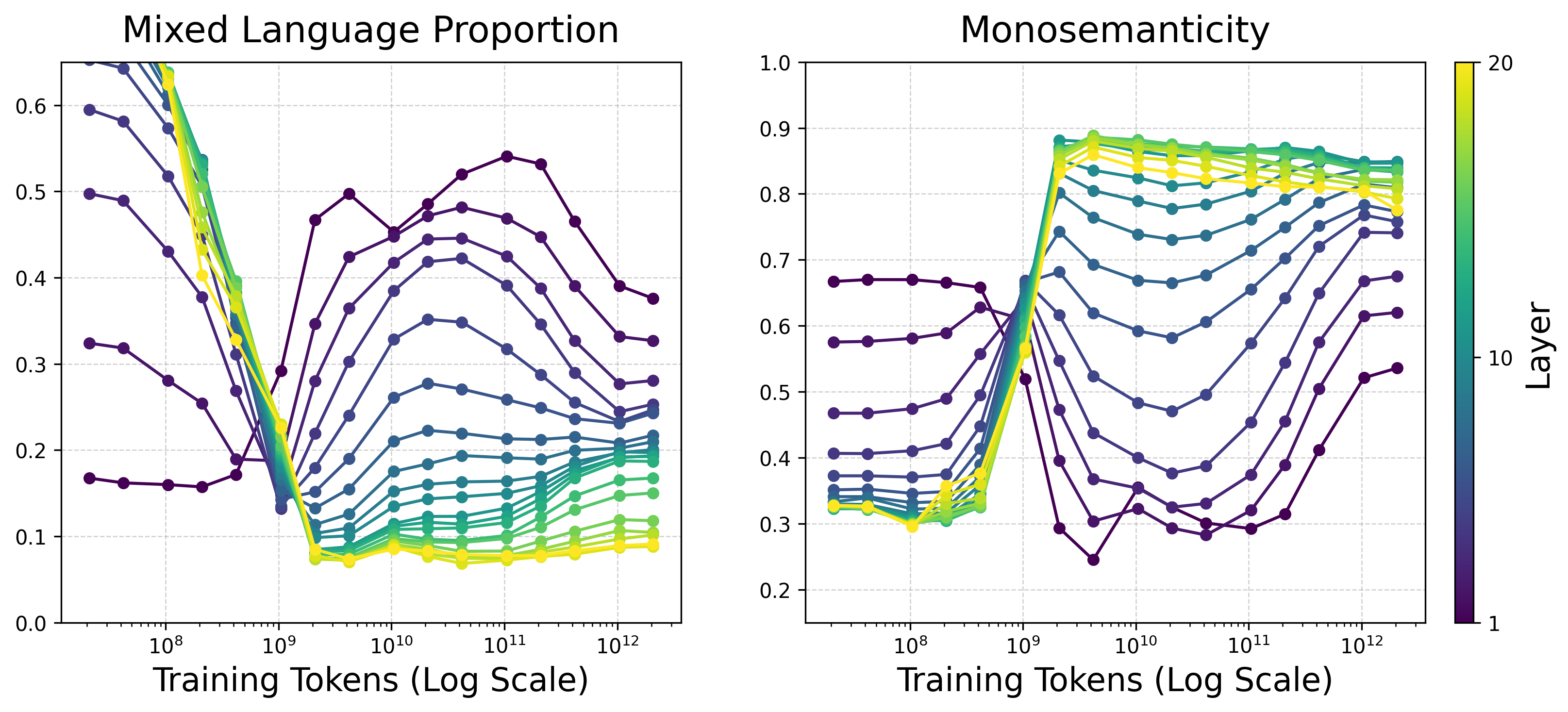}
    \caption{Layer-wise evolution of the mixed language proportion and the monosemanticity in the 980M model across training stages.}
    \label{fig:mix_mono_980m}
\end{figure*}

\begin{figure*}[tbp]
    \centering
    \includegraphics[width=\textwidth]{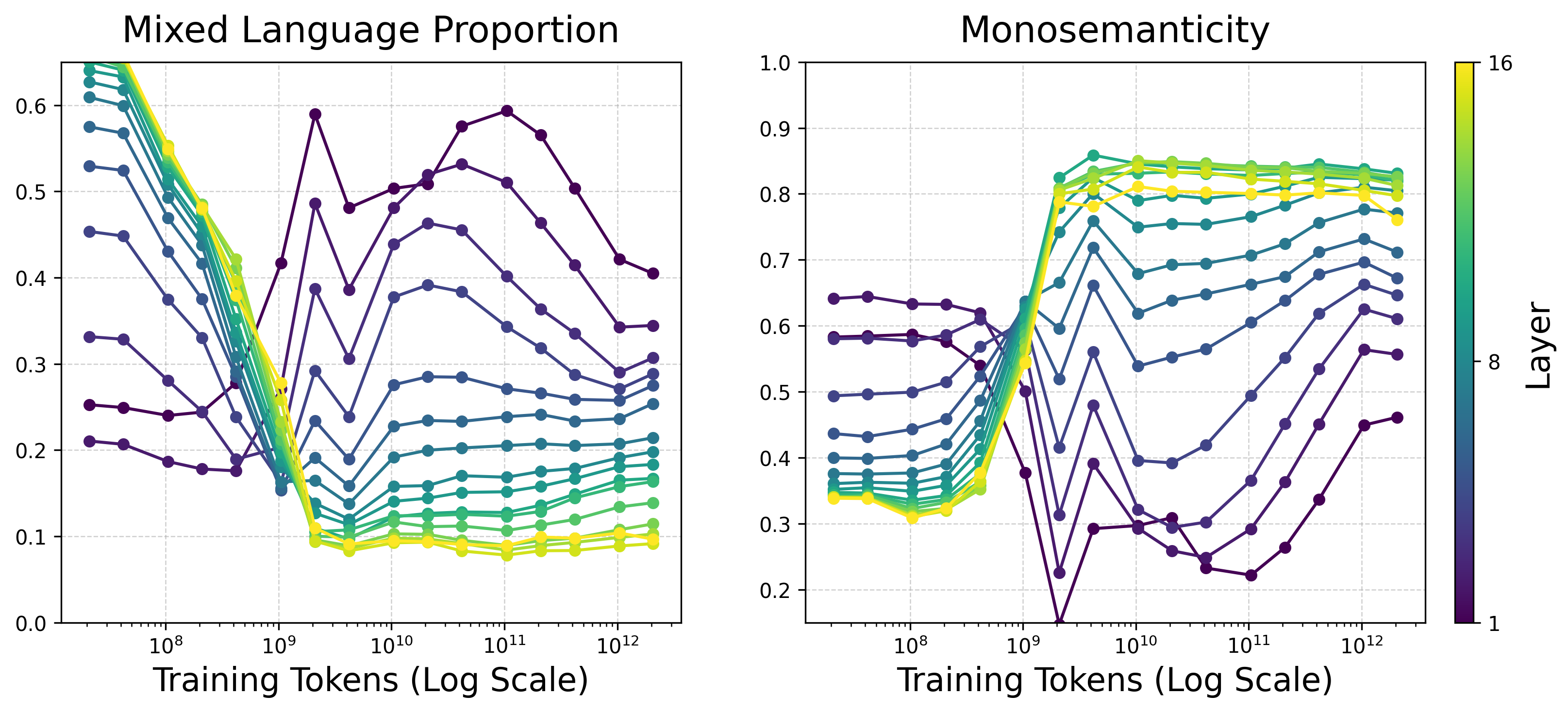}
    \caption{Layer-wise evolution of the mixed language proportion and the monosemanticity in the 440M model across training stages.}
    \label{fig:mix_mono_440m}
\end{figure*}

\begin{figure*}[tbp]
    \centering
    \includegraphics[width=\columnwidth]{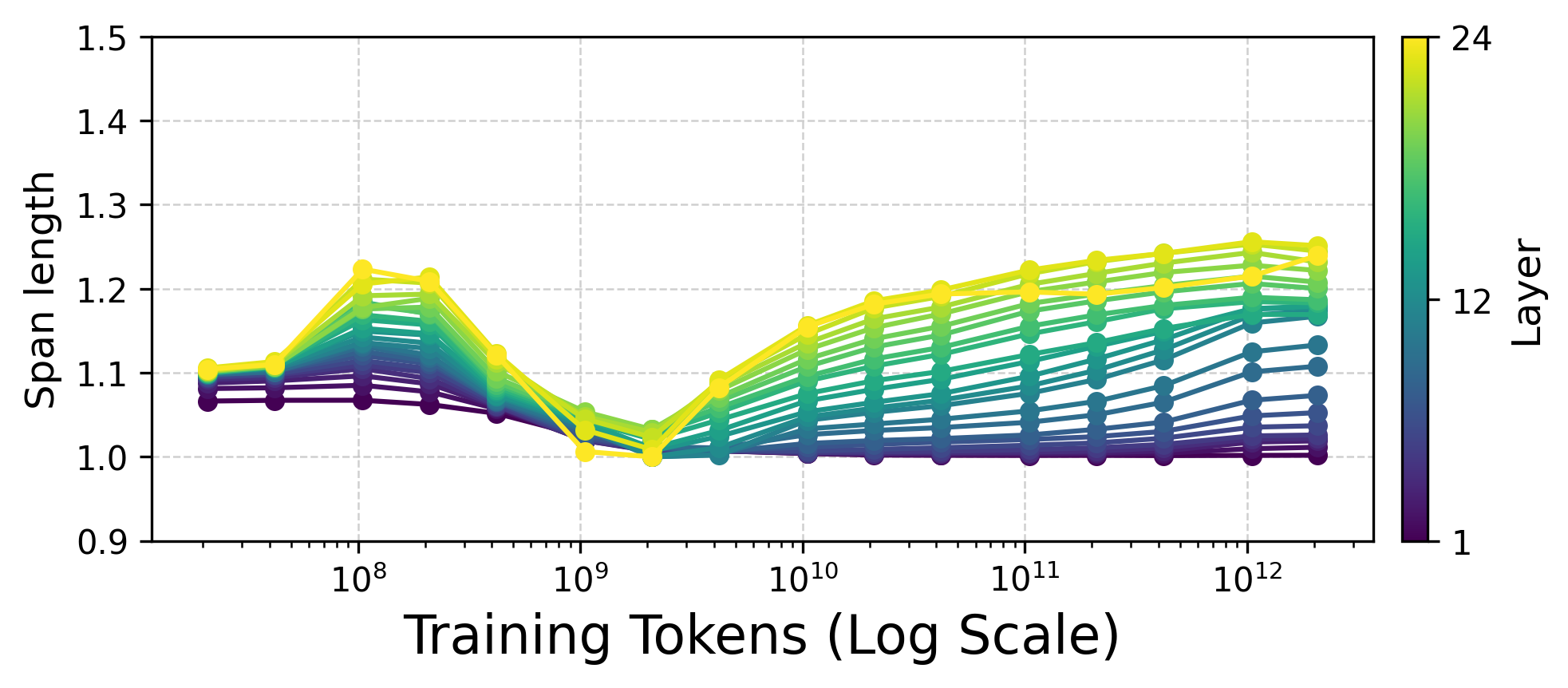}
    \caption{Layer-wise evolution of the span length average in the 1.8B model across training stages.}
    \label{fig:span_1.8b}
\end{figure*}
\begin{figure*}[tbp]
    \centering
    \includegraphics[width=\columnwidth]{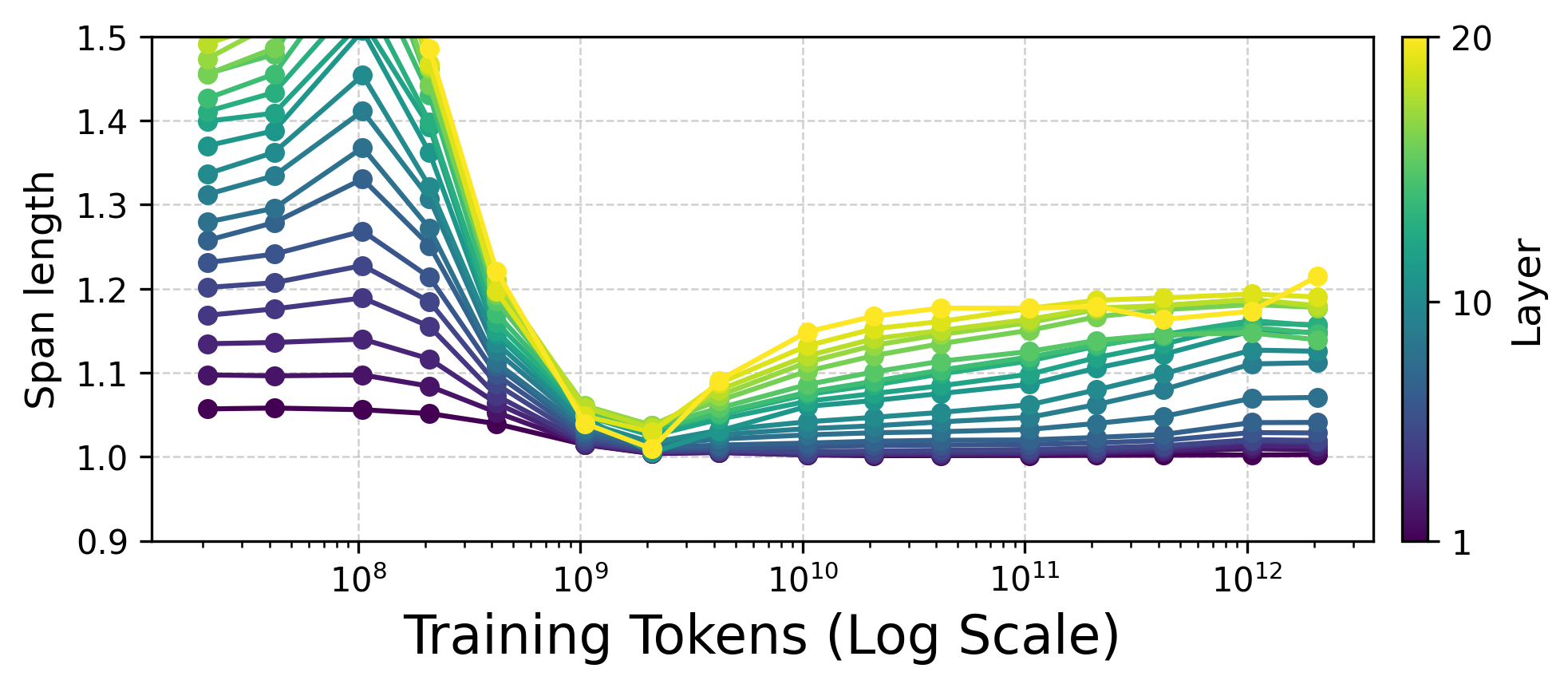}
    \caption{Layer-wise evolution of the span length average in the 980M model across training stages.}
    \label{fig:span_980m}
\end{figure*}
\begin{figure*}[tbp]
    \centering
    \includegraphics[width=\columnwidth]{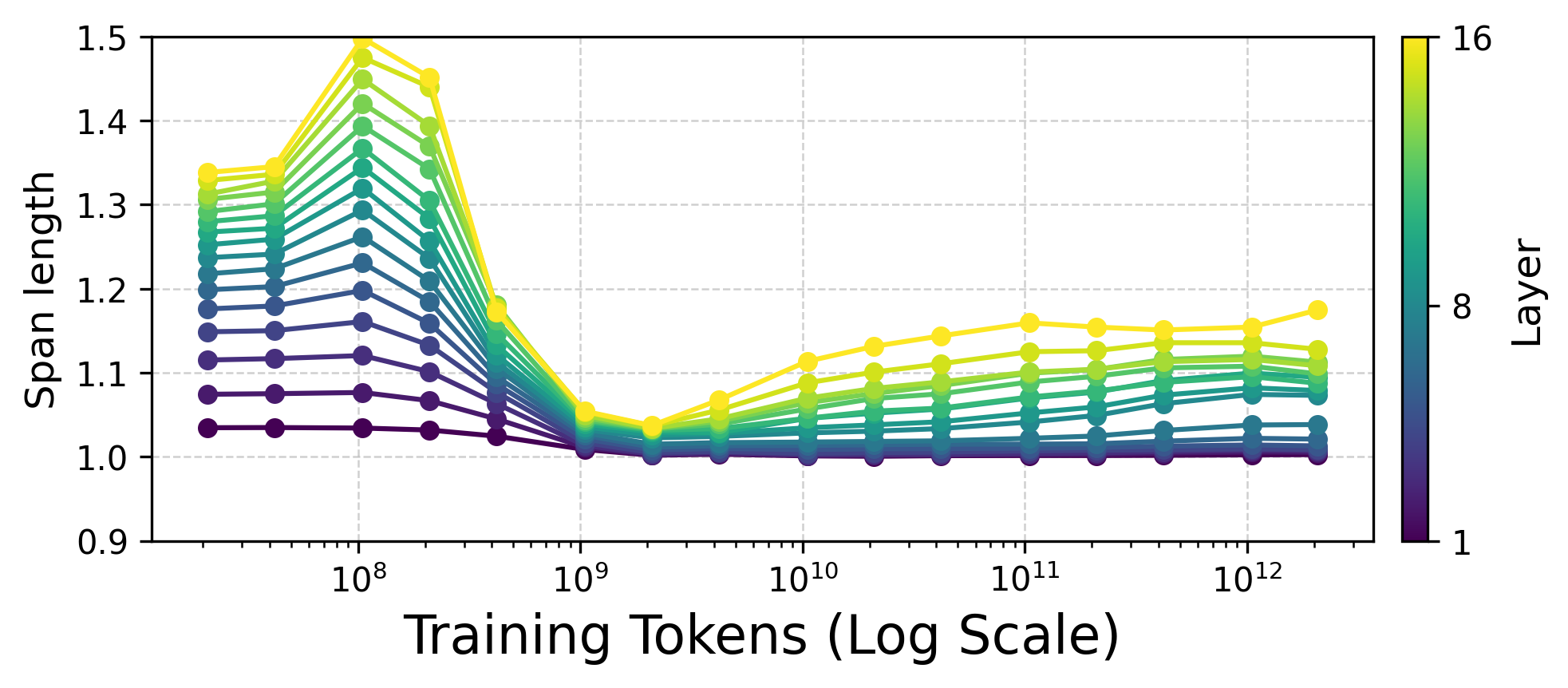}
    \caption{Layer-wise evolution of the span length average in the 440M model across training stages.}
    \label{fig:span_440m}
\end{figure*}

\end{document}